\documentclass{article}
\usepackage{spconf,amsmath,graphicx}
\usepackage{cite}
\usepackage{amsfonts,bm}
\usepackage{subfigure}
\usepackage{multicol,multirow}
\usepackage[normalem]{ulem}
\pdfoutput=1
\useunder{\uline}{\ul}{}

\title{FSOINet: Feature-Space Optimization-Inspired Network for Image Compressive Sensing}
\name{Wenjun Chen, Chunling Yang*, Xin Yang\thanks{Supported by the Key Program of Natural Science Foundation of Guangdong Province (2017A030311028) and the Natural Science Foundation of Guangdong Province (2019A1515011949).  * means corresponding author}}
\address{School of Electronic and Information Engineering, South China University of Technology, China\\(eecwjjun@mail.scut.edu.cn; eeclyang@scut.edu.cn; eexyang@mail.scut.edu.cn)}

\begin{document}

\maketitle

\begin{abstract}
	In recent years, deep learning-based image compressive sensing (ICS) methods have achieved brilliant success. Many optimization-inspired networks have been proposed to bring the insights of optimization algorithms into the network structure design and have achieved excellent reconstruction quality with low computational complexity. But they keep the information flow in pixel space as traditional algorithms by updating and transferring the image in pixel space, which does not fully use the information in the image features. In this paper, we propose the idea of achieving information flow phase by phase in feature space and design a Feature-Space Optimization-Inspired Network (dubbed FSOINet) to implement it by mapping both steps of proximal gradient descent algorithm from pixel space to feature space. Moreover, the sampling matrix is learned end-to-end with other network parameters. Experiments show that the proposed FSOINet outperforms the existing state-of-the-art methods by a large margin both quantitatively and qualitatively. The source code is available on https://github.com/cwjjun/FSOINet.
\end{abstract}

\begin{keywords}
	Image compressive sensing, deep learning, convolutional neural network, image reconstruction
\end{keywords}

\begin{figure*}[htb]
	\centering
	\centerline{\includegraphics[width=\textwidth]{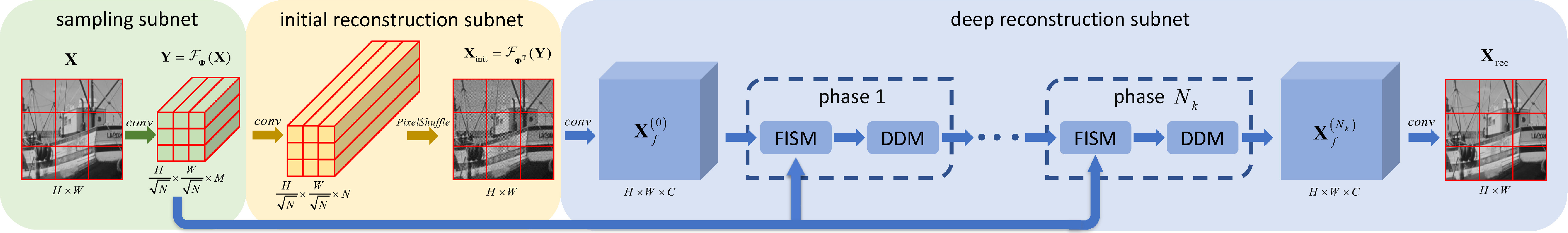}}
	\caption{The architecture of the FSOINet.}
	\label{fig:Net}
\end{figure*}

\section{Introduction}
\label{sec:intro}
Compressive sensing (CS)~\cite{cs}, which demonstrates that a signal can be reconstructed by fewer measurements than the Nyquist sampling theorem when it has sparsity in a proper transform domain, has gained increasing attention for several years. As an inverse problem, CS aim to reconstruct the original signal $\bm{x}\in\mathbb{R}^N$ from its CS measurements $\bm{y}\in \mathbb{R}^M(M\ll N)$ gained with a linear projection $\bm{y}=\Phi\bm{x}$, where $\Phi\in \mathbb{R}^{M \times N}$. The inverse problem is ill-posed and can be converted into the following optimization problem:
\begin{equation}\label{eq:optimization_goal}
	\mathop{\min}_{\bm{x}}\frac{1}{2}\Vert{{\Phi}\bm{x}-\bm{y}}\Vert^2_2+\lambda\psi(\bm{x}),
\end{equation}
where $\psi(\bm{x})$ comes from prior knowledge and $\lambda$ is a regularization parameter. Iterative optimization algorithms usually use proximal gradient descent to solve this problem, which can be depicted as two update steps:
\begin{equation}\label{eq:r_model}
	\bm{r}^{(k)}=\bm{x}^{(k-1)}-\rho\Phi^{\mathrm{T}}(\Phi \bm{x}^{(k-1)}-\bm{y}),
\end{equation}
\begin{equation}\label{eq:x_model}
	\bm{x}^{(k)}=\text{prox}_{\lambda,\psi}(\bm{r}^{(k)}),
\end{equation}
where $\text{prox}_{\lambda,\psi}(\bm{r})=\mathop{\arg\min}_{\bm{x}}\frac{1}{2}\Vert{\bm{x}}-\bm{r}\Vert^2_2+\lambda\psi(\bm{x})$, $k$ denotes the iteration index and $\rho$ is the step size.

It’s well known that natural image has lots of redundant information. In consequence, the study of how to use CS theory in image compression has become more significant. Because of the high resolution of the image, Block-based Compressive Sensing (BCS)~\cite{bcs} was proposed to sample the image more efficiently by sampling the image block by block. Recently, Convolution Neural Networks (CNN) have been successfully used in many computer vision tasks. First CNN-based ICS method~\cite{reconnet} provides improved performance in terms of reconstruction quality and speed than traditional iterative optimization-based CS methods. Subsequent CNN-based ICS methods could be divided into two categories, one is vanilla neural networks and the other is optimization-inspired neural networks. The former~\cite{lapran,csnet,scsnet,macnet} build neural networks as a black box to learn the non-linear mapping from the CS measurements to the reconstructed image. The latter~\cite{admmnet,istanet,splnet,ampnet,opinenet,coast} unfold iterative algorithms to neural networks comprising fixed numbers of phases, and each phase corresponds to an iteration in traditional algorithms.

Although optimization-inspired neural networks usually get better reconstruction quality by using measurements in each reconstruction phase, they still keep the information flow in pixel space between phases as traditional algorithms. Each phase supplements information in pixel space according to Eq.\ref{eq:r_model} and only transfers pixel-space image denoised by the network, which suffers from two limitations. First, the simple process of correcting information in pixel space from measurements can not fully exploit the information contained in measurements. Second, the structure which transfers information in pixel space doesn’t make full use of the information contained in the image features.

To overcome these drawbacks, we build a novel ICS network structure unfolding iterative optimization algorithms in feature space, dubbed Feature-Space Optimization-Inspired Network (FSOINet) for ICS, which processes and transfers image features phase by phase to make full use of features extracted from the image. In each phase, we use measurements to supplement the information in feature space and then denoise the image feature by building a Feature-space Information Supplementing Module (FSIM) and a Dual-scale Denoising Module (DDM).

The main contributions of this paper are summarized as follows: 1) We propose the idea of implementing the information flow phase by phase in feature space for ICS. 2) A FSIM and a DDM are designed to implement update steps in feature space for supplementing the information from measurements and dual-scale denoising, respectively. 3) A novel deep unfolding network named FSOINet is proposed, which contains a sampling subnet, an initial reconstruction subnet, and a deep reconstruction subnet. 4) Experiment shows our framework outperforms other state-of-the-art ICS methods significantly in terms of both reconstruction quality and visual quality.

\section{The Proposed FSOINet}
\label{sec:prop}
Inspired by the iterative optimization algorithm, as shown in Fig.\ref{fig:Net}, we built FSOINet comprising three subnets: a sampling subnet, an initial reconstruction subnet, and a deep reconstruction subnet. The first two subnets respectively realize the linear sampling and initial reconstruction between the pixel domain and the measurement domain. The last subnet completes the non-linear reconstruction of the image by processing image features, which contains $N_k$ phases and each of them corresponds to an iteration in the iterative process.

\subsection{Sampling and Initial Reconstruction}
\label{ssec:samp_init}
In the BCS~\cite{bcs} strategy, an image $\bm{X}\in\mathbb{R}^{H \times W}$ is divided into $\frac{H}{\sqrt N}\times \frac{W}{\sqrt N}$ non-overlapping image blocks with size of $\sqrt N  \times \sqrt N $, then each of them is reshaped into a vector $\bm{x}\in\mathbb{R}^N$. When the CS sampling rate is $r$, the same sampling matrix $\Phi\in \mathbb{R}^{M \times N}$ is used to sample each $\bm{x}$, where $M = \left\lfloor r \times N \right\rfloor$. In addition, Cui et al.~\cite{dsmm} proposes to use a learnable sampling matrix instead of fixed random Gaussian Matrix to get CS measurements, and have achieved gratifying results. In this paper, we use a sampling sub-network which is denoted as $\mathcal{F}_\Phi(\cdot)$  to sample the image and use convolution without bias to simulate the block sampling process described above. To be specific, we set $\Phi$ as learnable network parameters, reshape each row of $\Phi$ into a convolution kernel of size $1 \times \sqrt N  \times \sqrt N$ to obtain $W_\Phi$. Therefore, the sampling process can be described by convolution with a stride of $\sqrt N$ as:
\begin{equation}\label{eq:sampling}
	\bm{Y}=\mathcal{F}_\Phi(\bm{X})=W_\Phi\ast \bm{X}.
\end{equation}

To obtain a reasonable initial estimate of each image block from CS measurements and not introduce more learnable parameters in the meanwhile, we use $\Phi^{\mathrm{T}}$ to complete the initial reconstruction. Like the sampling process, each row of $\Phi^{\mathrm{T}}$ is reshaped into a convolution kernel of size $M \times 1  \times 1$ to obtain $W_\Phi^{\mathrm{T}}$, then PixelShuffle is used to obtain the initial reconstruction image, which can be illustrated as:
\begin{equation}\label{eq:initialing}
	\bm{X}_\text{init}=\mathcal{F}_{\Phi^{\mathrm{T}}}(\bm{X})=\text{PixelShuffle}(W_{\Phi^{\mathrm{T}}} \ast \bm{X}).
\end{equation}

\begin{figure}[tb]
	\begin{minipage}[b]{1\linewidth}
		\centering
		\centerline{\includegraphics[width=0.98\linewidth]{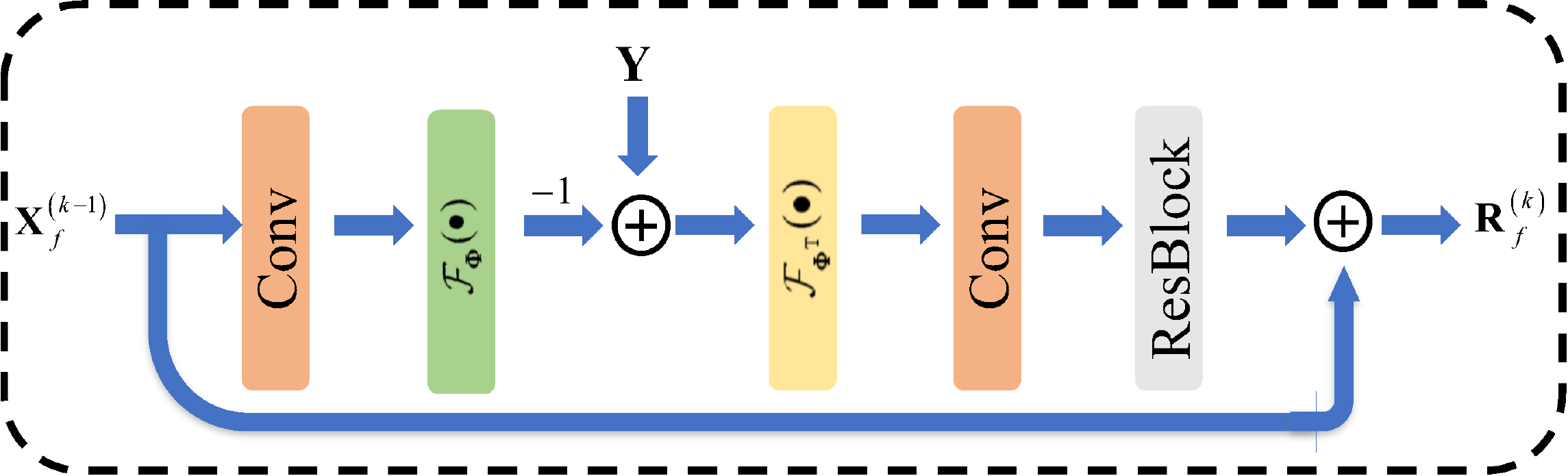}}
		\centerline{(a) Feature-space Information Supplementing Module}\medskip
	\end{minipage}
	\vfill
	\begin{minipage}[b]{.72\linewidth}
		\centering
		\centerline{\includegraphics[width=0.96\linewidth]{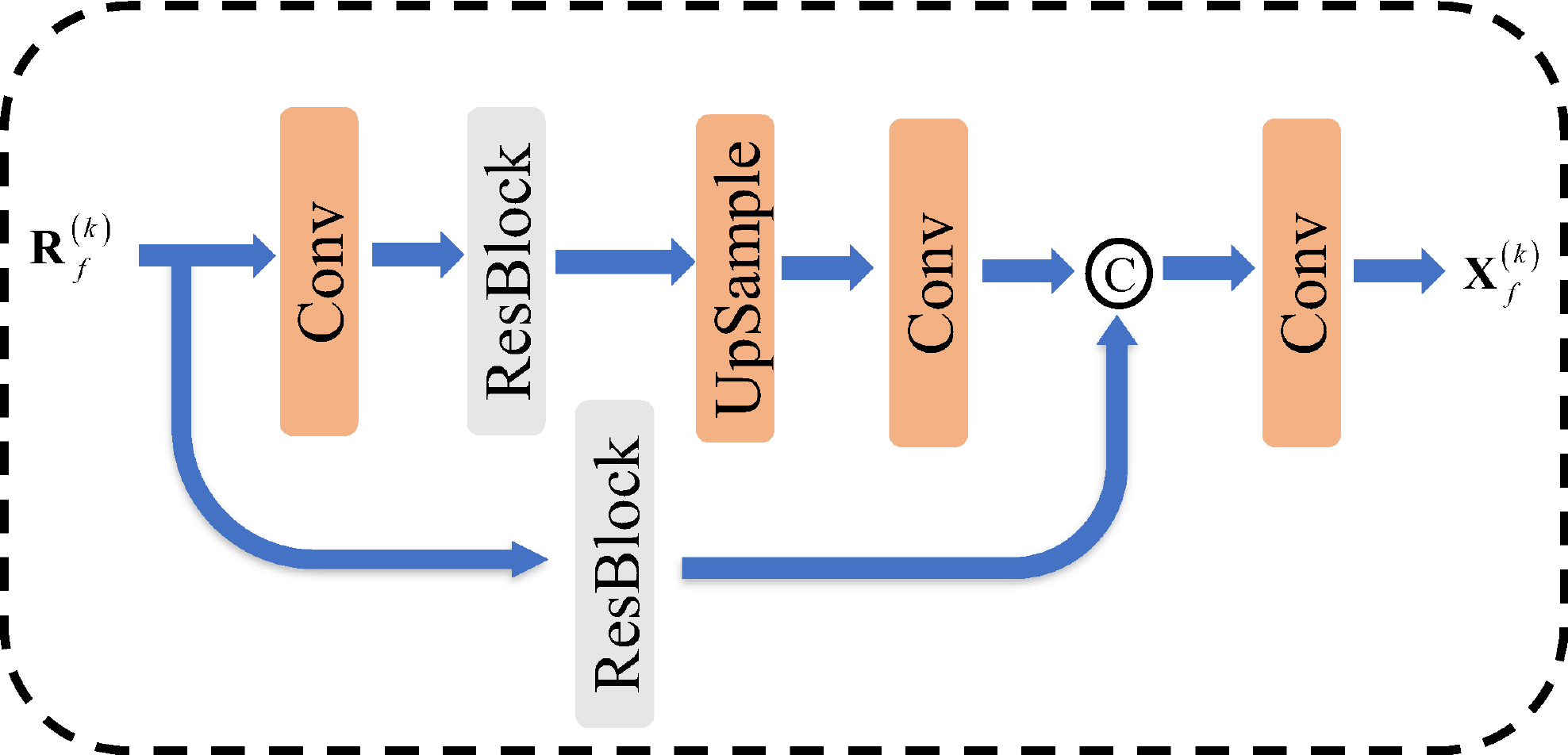}}
		\centerline{(b) Dual-scale Denoising Module}\medskip
	\end{minipage}
	\hspace{6mm}
	\begin{minipage}[b]{0.17\linewidth}
		\centering
		\centerline{\includegraphics[width=\linewidth]{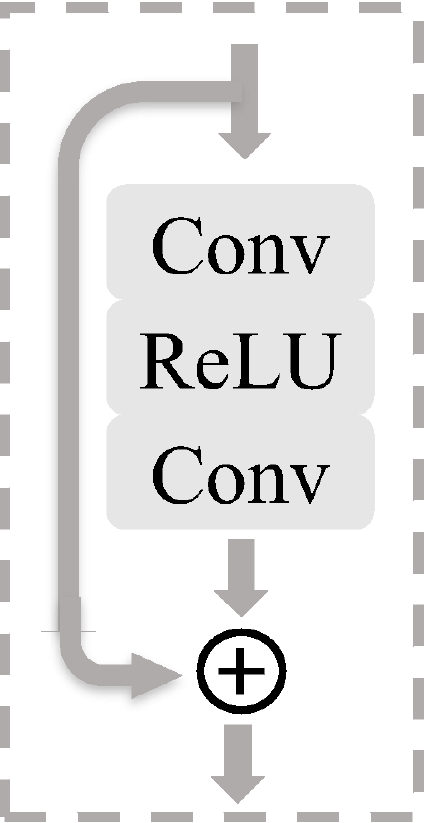}}
		\centerline{(c) ResBlock}\medskip
	\end{minipage}
	\caption{The architecture of the FSIM and DDM.}
	\label{fig:GEM_DDM}
\end{figure}

\begin{table*}[htp]
	\centering
	\caption{Average PSNR and SSIM comparisons of different ICS methods on Set11, BSD68 and BSD100.}
	\label{tab:result}
	\resizebox{0.83\textwidth}{!}{
		\begin{tabular}{c|l|lllll}
			\hline
			\multicolumn{1}{l|}{\multirow{2}{*}{Datasets}} & \multirow{2}{*}{methods} & \multicolumn{5}{c}{cs ratio r}                                                                                                                                                                                  \\ \cline{3-7}
			\multicolumn{1}{l|}{}                          &                          & \multicolumn{1}{c|}{0.01}                   & \multicolumn{1}{c|}{0.05}                   & \multicolumn{1}{c|}{0.1}                    & \multicolumn{1}{c|}{0.3}                    & \multicolumn{1}{c}{0.5} \\ \hline
			\multirow{6}{*}{Set11}                         & CSNet$^+$                & \multicolumn{1}{l|}{21.02/\uline{0.5566}}   & \multicolumn{1}{l|}{25.86/0.7846}           & \multicolumn{1}{l|}{28.34/0.8508}           & \multicolumn{1}{l|}{34.30/0.9490}           & 38.52/0.9749            \\
			                                               & SCSNet                   & \multicolumn{1}{l|}{21.04/0.5562}           & \multicolumn{1}{l|}{25.85/0.7839}           & \multicolumn{1}{l|}{28.52/0.8616}           & \multicolumn{1}{l|}{34.64/0.9511}           & 39.01/0.9769            \\
			                                               & SPLNet                   & \multicolumn{1}{l|}{{\uline{21.22}/0.5552}} & \multicolumn{1}{l|}{{\uline{26.59}/0.8177}} & \multicolumn{1}{l|}{29.49/0.8874}           & \multicolumn{1}{l|}{35.79/0.9603}           & 40.27/0.9815            \\
			                                               & AMP-Net                  & \multicolumn{1}{l|}{20.20/0.5425}           & \multicolumn{1}{l|}{26.17/0.8128}           & \multicolumn{1}{l|}{29.40/0.8876}           & \multicolumn{1}{l|}{36.03/\uline{0.9623}}   & {\ul 40.34/0.9821}      \\
			                                               & OPINE-Net$^+$            & \multicolumn{1}{l|}{20.02/0.5362}           & \multicolumn{1}{l|}{{26.36/\uline{0.8186}}} & \multicolumn{1}{l|}{{\ul 29.81/0.8904}}     & \multicolumn{1}{l|}{\uline{36.04}/0.9600}   & 40.19/0.9800            \\
			                                               & FSOINet                  & \multicolumn{1}{l|}{\textbf{21.73/0.5937}}  & \multicolumn{1}{l|}{\textbf{27.36/0.8415}}  & \multicolumn{1}{l|}{\textbf{30.44/0.9018}}  & \multicolumn{1}{l|}{\textbf{37.00/0.9665}}  & \textbf{41.08/0.9832}   \\ \hline
			\multirow{6}{*}{BSD68}                         & CSNet$^+$                & \multicolumn{1}{l|}{21.71/0.5249}           & \multicolumn{1}{l|}{25.04/0.6845}           & \multicolumn{1}{l|}{26.89/0.7756}           & \multicolumn{1}{l|}{31.66/0.9152}           & 35.42/0.9614            \\
			                                               & SCSNet                   & \multicolumn{1}{l|}{21.88/0.5250}           & \multicolumn{1}{l|}{24.98/0.6843}           & \multicolumn{1}{l|}{27.13/0.7785}           & \multicolumn{1}{l|}{31.76/0.9173}           & 35.67/0.9640            \\
			                                               & SPLNet                   & \multicolumn{1}{l|}{{\uline{22.33}/0.5242}} & \multicolumn{1}{l|}{{\uline{25.87}/0.7198}} & \multicolumn{1}{l|}{{\uline{27.85}/0.8094}} & \multicolumn{1}{l|}{32.77/0.9303}           & {\uline{36.86}/0.9708}  \\
			                                               & AMP-Net                  & \multicolumn{1}{l|}{{22.28/\uline{0.5315}}} & \multicolumn{1}{l|}{{25.77/\uline{0.7204}}} & \multicolumn{1}{l|}{{\ul 27.85/0.8113}}     & \multicolumn{1}{l|}{\ul 32.84/0.9321}       & {36.82/\uline{0.9715}}  \\
			                                               & OPINE-Net$^+$            & \multicolumn{1}{l|}{21.88/0.5162}           & \multicolumn{1}{l|}{25.66/0.7136}           & \multicolumn{1}{l|}{27.81/0.8040}           & \multicolumn{1}{l|}{32.50/0.9236}           & 36.32/0.9658            \\
			                                               & FSOINet                  & \multicolumn{1}{l|}{\textbf{22.75/0.5418}}  & \multicolumn{1}{l|}{\textbf{26.21/0.7324}}  & \multicolumn{1}{l|}{\textbf{28.27/0.8187}}  & \multicolumn{1}{l|}{\textbf{33.29/0.9348}}  & \textbf{37.34/0.9727}   \\ \hline
			\multirow{6}{*}{Urban100}                      & CSNet$^+$                & \multicolumn{1}{l|}{19.27/0.4812}           & \multicolumn{1}{l|}{22.63/0.6792}           & \multicolumn{1}{l|}{24.64/0.7741}           & \multicolumn{1}{l|}{29.90/0.9162}           & 33.55/0.9572            \\
			                                               & SCSNet                   & \multicolumn{1}{l|}{19.28/0.4798}           & \multicolumn{1}{l|}{22.63/0.6774}           & \multicolumn{1}{l|}{24.93/0.7827}           & \multicolumn{1}{l|}{30.12/0.9193}           & 33.92/0.9601            \\
			                                               & SPLNet                   & \multicolumn{1}{l|}{19.55/0.4873}           & \multicolumn{1}{l|}{23.55/0.7301}           & \multicolumn{1}{l|}{26.19/0.8290}           & \multicolumn{1}{l|}{32.11/0.9405}           & {36.41/\uline{0.9737}}  \\
			                                               & AMP-Net                  & \multicolumn{1}{l|}{{\ul 19.62/0.4969}}     & \multicolumn{1}{l|}{23.45/0.7290}           & \multicolumn{1}{l|}{26.04/0.8283}           & \multicolumn{1}{l|}{32.19/\uline{0.9418}}   & {36.33/\uline{0.9737}}  \\
			                                               & OPINE-Net$^+$            & \multicolumn{1}{l|}{19.38/0.4872}           & \multicolumn{1}{l|}{{\ul 23.70/0.7363}}     & \multicolumn{1}{l|}{{\ul 26.61/0.8362}}     & \multicolumn{1}{l|}{{\uline{32.58}/0.9414}} & {\uline{36.62}/0.9727}  \\
			                                               & FSOINet                  & \multicolumn{1}{l|}{\textbf{19.87/0.5223}}  & \multicolumn{1}{l|}{\textbf{24.57/0.7750}}  & \multicolumn{1}{l|}{\textbf{27.53/0.8627}}  & \multicolumn{1}{l|}{\textbf{33.84/0.9540}}  & \textbf{37.80/0.9777}   \\ \hline
		\end{tabular}
	}
\end{table*}

\subsection{Deep Reconstruction}
\label{ssec:recon}
In previous optimization-inspired ICS networks, each phase implements supplementing information in pixel space according to Eq.\ref{eq:r_model} and sends the denoised image to the next phase, although the denoising is finished in feature space. This process makes the information flow in pixel space phase by phase, which limits the use of the robust feature representation ability of CNN. In this paper, we propose the idea of implementing the information flow phase by phase in feature space, including supplementing the information from measurements and dual-scale denoising all in the feature space, to fully utilize the feature representation ability of CNN and retain more details when the information flows to the next phase. The deep reconstruction subnet is shown in Fig.\ref{fig:Net}.

\textbf{Information supplement in feature space:} Consistent with the optimization algorithm, we also use measurements throughout the entire reconstruction process to guarantee the solution complies with the degradation process $\bm{y}=\Phi\bm{x}$. And different from previous optimization-inspired ICS methods, we propose the idea of supplementing information in feature space instead of that in pixel space from measurements. To implement this idea, we build a Feature-space Information Supplementing Module (FSIM) to map the gradient of the fidelity term $\frac{1}{2}\Vert{{\Phi}\bm{x}-\bm{y}}\Vert^2_2$ in Eq.\ref{eq:optimization_goal} to the feature space and fuse it with the image features, which is shown in Fig.\ref{fig:GEM_DDM}(a). It is worth noting that $\mathcal{F}_\Phi(\cdot)$ and $\mathcal{F}_\Phi^{\mathrm{T}}(\cdot)$  in Fig.\ref{fig:GEM_DDM}(a) are block operations, which will introduce blocking artifacts. Here, ResBlock operates on the entire image, which can suppress blocking artifacts.

\textbf{Dual-scale denoising in feature space:} Benefiting from the data-driven characteristics of deep learning, the inherent prior knowledge of the image can be learned from the training set to denoise the image. In this paper, we propose a dual-scale structure network to improve denoising efficiency. And we design a neural network module called Dual-scale Denoising Module (DDM) to implement it, as shown in Fig.\ref{fig:GEM_DDM}(b). To maintain not too many parameters, only two ResBlocks are used to denoise the high-resolution features and low-resolution features respectively in DDM. Then, the denoised features are fused to obtain the output features, which will be sent into the next phase. Finally, we implement the information flow in feature space phase by phase. The high-resolution features maintain the same resolution as the original image, while the low-resolution features are subsampled from the high-resolution features with stride-2 Conv and have twice the number of channels.

\subsection{Loss Function}
\label{ssec:loss}
For an original image $\bm{X}$, the proposed model first obtains CS measurements $\bm{Y}$ by sampling it and then predicts the reconstructed image $\bm{X}_\text{rec}$. We optimize our FSOINet end-to-end through the following loss function:
\begin{equation}\label{eq:loss}
	\mathcal{L}=\mathcal{L}_{mse}(\bm{X}_\text{rec},\bm{X})+\gamma\mathcal{L}_{orth}(\Phi),
\end{equation}
where $\mathcal{L}_{mse}$ is the Mean Squared Error (MSE) of the original image $\bm{X}$ and the reconstructed image ${X}_\text{rec}$. In addition, $\mathcal{L}_{orth}$ is the orthogonal constraint of the sampling matrix in \cite{opinenet}:$\mathcal{L}_{orth}(\Phi)=\frac{1}{M^2}\Vert{{\Phi\Phi^{\mathrm{T}}}-\text{I}}\Vert^2_F$, where \text{I} represents the identity matrix. The regularization parameter $\gamma$ in Eq.\ref{eq:loss} is set to 0.01 in our experiments.

\begin{figure*}[htb]
	\centering
	\centerline{\includegraphics[width=0.82\textwidth]{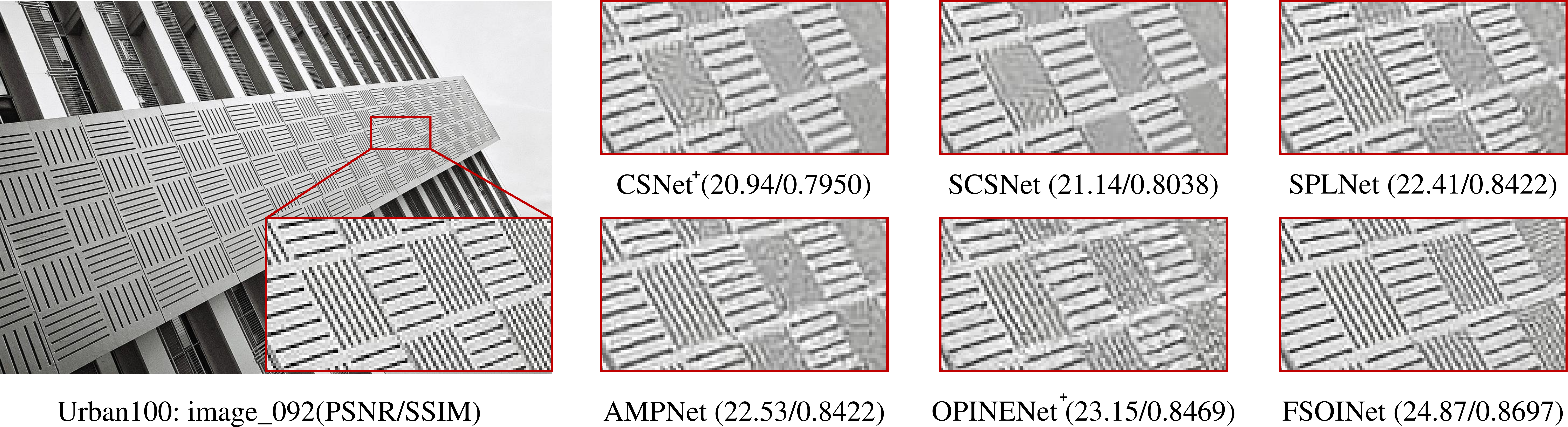}}
	\caption{Visual comparison of various ICS methods on one sample from Urban100 at the CS sampling ratio of 0.3.}
	\label{fig:result}
\end{figure*}

\section{Experiment Result}
\label{sec:result}
\subsection{Dataset and Implementation Details}
\label{ssec:dataset}
For training, following~\cite{csnet}, we use 400 images from the training set and test set of the BSDS500 dataset~\cite{bsds500}. The training images are cropped to 89600 $96\times96$ pixel sub-images with data augmentation. For the network parameters, the block size $\sqrt N$ is 32, the channel number $C$ is 16, the phase number $N_k$ is 16 and the batch size is 32. The size of the unspecified convolution kernel is $3 \times 3 $. We use Adam~\cite{adam} optimizer to train the network with the initial learning rate of $2\times10^{- 4}$, which is decreased to $5\times10^{- 5}$ through 100 epochs using the cosine annealing strategy~\cite{coslr} and the warm-up epochs are 3. All the experiments are implemented in PyTorch with a Core i5-6500 CPU Intel and a GTX2080ti GPU. For testing, we utilize three widely-used benchmark datasets, including Set11~\cite{reconnet}, BSDS68~\cite{bsd68} and Urban100~\cite{urban100}. Color images are processed in the YCbCr space and evaluated on the Y channel. Two common-used image assessment criteria, Peak Signal to Noise Ratio (PSNR) and Structural Similarity (SSIM), are adopted to evaluate the reconstruction results.

\begin{table}[tp]
	\centering
	\caption{Model size and average computational time of different ICS methods on BSD68 (r = 0.5).}
	\label{tab:complexity}
	\resizebox{0.48\textwidth}{!}{
		\begin{tabular}{ccccc}
			\hline
			Model  & SPLNet  & AMPNet  & OPINENet$^+$ & FSOINet \\ \hline
			\#Para & 1.39M   & 1.53M   & 1.10M        & 1.06M   \\ \hline
			Time   & 0.0059s & 0.0466s & 0.0109s      & 0.0239s \\ \hline
		\end{tabular}
	}
\end{table}

\subsection{Comparisons with State-of-the-Arts}
\label{ssec:compare}
Considering that the sampling matrix obtained by training is often better than the fixed Gaussian random matrix. In order to make fair comparisons, we compare our FSOINet with five state-of-the-art ICS methods whose sampling matrixes are learnable too, including CSNet~\cite{csnet}, SCSNet~\cite{scsnet}, SPLNet~\cite{splnet}, AMPNet~\cite{ampnet}, and OPINE-Net~\cite{opinenet}. Extensive experiments have proved the advantages of FSOINet in terms of quality and visualized results.

Table \ref{tab:result} and Fig.\ref{fig:result} clearly show that our FSOINet outperforms all the other competing methods at all the CS sampling rates. It is worth noting that the three optimization-inspired methods, SPLNet, OPINENet, and AMPNet, are significantly better than vanilla neural networks CSNet and SCSNet in terms of reconstruction quality. But our FSOINet got a more compelling reconstruction quality and reconstruct sharper edges and clearer background at all the sampling rates.

Table \ref{tab:complexity} provides a comparison of model sizes and calculation times for different methods at a CS ratio of 0.5. Our model has the fewest parameters and medium calculation speed while achieving the best reconstruction results.

\subsection{Ablation Study}
\label{ssec:abla}
This section mainly analyzes the effectiveness of using FSIM to replace the gradient descent in the optimization algorithm. We retrain our network without FSIM, then the network is similar to vanilla neural networks, which only use the measurements during the initial reconstruction, dubbed VNet. At the same time, we change the FSIM to the gradient descent in the pixel domain and retrain our network. At this time, the network which we named OINet is more similar to the previous optimization-inspired neural networks. For a fair comparison, we move the convolution block used in FSIM to DDM to maintain similar model parameters.

As shown in Table \ref{tab:ablation}, the PSNR/SSIM of OINet is 0.54dB/0.0076 lower than that of FSOINet at the sampling ratio of 0.1, which reflects the effectiveness of mapping the gradient information to the feature space. While VNet’s PSNR/SSIM is 0.57db/0.0047 lower than that of OINet, which indicates the insights of the optimization algorithm could help the neural network structure design.

\begin{table}[tb]
	\centering
	\caption{Analysis of FSIM in FSOINet. The experiment are evaluated on Set11 (r = 0.1).}
	\label{tab:ablation}
	\resizebox{0.48\textwidth}{!}{
		\begin{tabular}{cccc}
			\hline
			Model      & VNet         & OINet        & FSOINet      \\ \hline
			PSNR/SSIM  & 29.33/0.8895 & 29.90/0.8942 & 30.44/0.9018 \\ \hline
			Parameters & 536257       & 536080       & 536257       \\ \hline
		\end{tabular}}
\end{table}

\section{Conclusion}
\label{sec:conc}
In this paper, we propose a novel ICS network structure dubbed FSOINet. The optimization algorithm is unfolded in the feature space through FSIM and DDM so that the information flow is kept phase by phase in the feature space. The experimental results show our model has achieved a significant improvement in performance compared with other state-of-the-art methods.

\vfill\pagebreak
\bibliographystyle{IEEEbib}
\bibliography{strings,refs}

\end{document}